\documentclass[10pt, a4paper]{article}
\usepackage[final]{lrec2026} %

\usepackage{booktabs} %
\usepackage[normalem]{ulem} %
\useunder{\uline}{\ul}{} %
\usepackage{array}

\usepackage{kotex} %
\ifxetexorluatex %
\setmainhangulfont{NanumBarunGothic.otf}[ BoldFont = NanumBarunGothicBold.otf , Path = {./font/} ]
\setmainhanjafont{NanumBarunGothic.otf}[ BoldFont = NanumBarunGothicBold.otf , Path = {./font/} ]
\newfontfamily{\hhfontyet}{NanumBarunGothic-YetHangul.otf}[ Path = {./font/} ]
\NewDocumentCommand{\fontyet}{m o}{{\hhfontyet #1}}
\else
\NewDocumentCommand{\fontyet}{m o}{\IfValueTF{#2}{#2}{Not rendered in arXiv}}
\fi

\usepackage{cleveref}
\usepackage{graphicx}
\usepackage{multirow}
\usepackage{tabularx}
\usepackage{xspace}
\usepackage{amssymb}
\usepackage{pifont}
\usepackage{listings}

\graphicspath{ {./images/} }
\crefformat{section}{\S#2#1#3}
\crefformat{subsection}{\S#2#1#3}
\crefformat{subsubsection}{\S#2#1#3}

\def\eg{\emph{e.g}.,\xspace}

\newcommand{\cmark}{\textcolor[HTML]{00AA00}{\ding{51}}}%
\newcommand{\xmark}{\textcolor[HTML]{CC0000}{\ding{55}}}%
\newcommand{\tmark}{\textcolor[HTML]{FFCC33}{\ding{115}}}

\newif\ifplaceholders
\placeholderstrue  %
\newcount\myloopcounter
\newcommand{\repeatit}[2][10]{%
  \ifplaceholders
  \myloopcounter0%
  \loop\ifnum\myloopcounter < #1%
  #2%
  \advance\myloopcounter by 1%
  \repeat%
  \fi
}

\newcommand{\corpus}{Open Korean Historical Corpus\xspace}
\title{\corpus{}: \\A Millennia-Scale Diachronic Collection of Public Domain Texts}

\name{
  Seyoung Song$^1$
  \hspace{4mm}
  Nawon Kim$^2$
  \hspace{4mm}
  Songeun Chae$^1$
  \hspace{4mm}
  Kiwoong Park$^1$
  \\
  {\bfseries \large Jiho Jin$^1$}
  \hspace{4mm}
  {\bfseries \large Haneul Yoo$^1$}
  \hspace{4mm}
  {\bfseries \large Kyunghyun Cho$^{3}$}
  \hspace{4mm}
  {\bfseries \large Alice Oh$^1$}
  \vspace{0.5mm}
}
\address{
  $^{1}$KAIST
  \hspace{4mm}
  $^{2}$Korea University
  \hspace{4mm}
  $^{3}$New York University
  \\
  \texttt{\href{mailto:seyoung.song@kaist.ac.kr}{\color{black}{seyoung.song@kaist.ac.kr}}, \href{mailto:knawon08@korea.ac.kr}{\color{black}{knawon08@korea.ac.kr}},}
  \\
  \texttt{\{\href{mailto:songeun@kaist.ac.kr}{\color{black}{songeun}}, \href{mailto:marspak@kaist.ac.kr}{\color{black}{marspak}}, \href{mailto:jinjh0123@kaist.ac.kr}{\color{black}{jinjh0123}}, \href{mailto:haneul.yoo@kaist.ac.kr}{\color{black}{haneul.yoo}}\}@kaist.ac.kr,}
  \\
  \texttt{\href{mailto:kyunghyun.cho@nyu.edu}{\color{black}{kyunghyun.cho@nyu.edu}}, \href{mailto:alice.oh@kaist.edu}{\color{black}{alice.oh@kaist.edu}}}
  \\
}

\abstract{
  The history of the Korean language is characterized by a discrepancy between its spoken and written forms and a pivotal shift from Chinese characters to the Hangul alphabet.
However, this linguistic evolution has remained largely unexplored in NLP due to a lack of accessible historical corpora.
To address this gap, we introduce the \corpus{}, a large-scale, openly licensed dataset spanning 1,300 years and 6 languages, as well as under-represented writing systems like Korean-style Sinitic (Idu) and Hanja-Hangul mixed script.
This corpus contains 17.7 million documents and 5.1 billion tokens from 19 sources, ranging from the 7th century to 2025.
We leverage this resource to quantitatively analyze major linguistic shifts: (1) Idu usage peaked in the 1860s before declining sharply; (2) the transition from Hanja to Hangul was a rapid transformation starting around 1890; and (3) North Korea's lexical divergence causes modern tokenizers to produce up to 51 times higher out-of-vocabulary rates.
This work provides a foundational resource for quantitative diachronic analysis by capturing the history of the Korean language.
Moreover, it can serve as a pre-training corpus for large language models, potentially improving their understanding of Sino-Korean vocabulary in modern Hangul as well as archaic writing systems.
 \\ \newline
}

\begin{document}

\maketitleabstract

\section{Introduction}
\label{sec:introduction}
Written communication on the Korean peninsula before the 20th century is linguistically compelling due to the inherent discrepancy between its spoken language and writing systems.
Prior to the creation of the native alphabet, Hangul (한글), Koreans borrowed Hanja (漢字) and used Classical Chinese as the primary literary language.
Because the grammatical structure of Korean (SOV) differs from that of Classical Chinese (SVO), unique transcription systems like Idu were developed to better reflect native grammar~\cite{handel2019sinography}.
Even after the invention of Hangul in the 15th century, a centuries-long transition involving mixed script usage preceded the predominantly Hangul-based writing of today.
The very success of this transition, however, has rendered centuries of Hanja-based records illegible to most contemporary Koreans, making its accessibility a significant challenge for modern NLP.

\begin{figure}[ht]
  \centering
  \includegraphics[width=\linewidth]{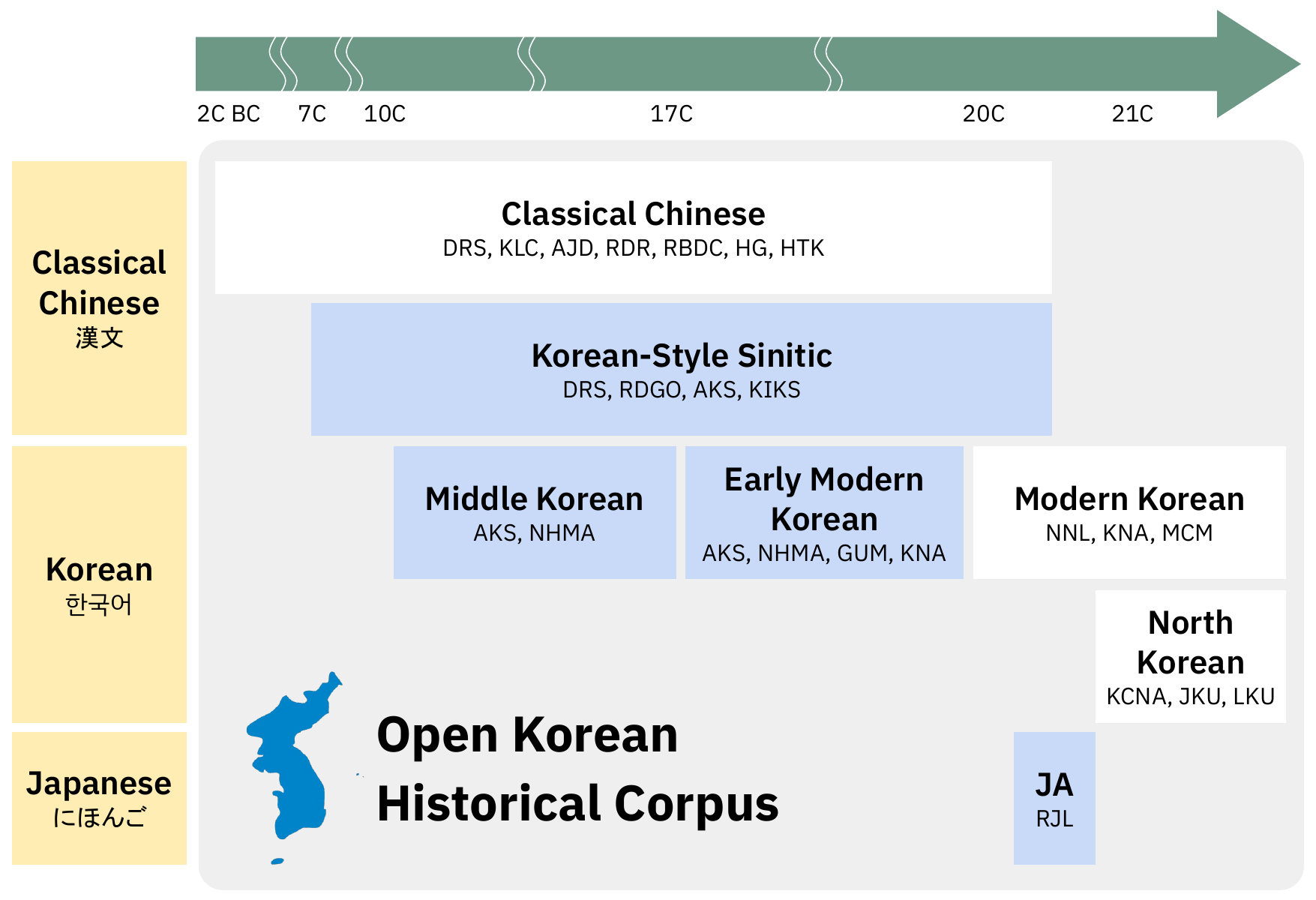}
  \caption{Overview of \corpus{}, which covers major languages and distinct writing systems used on the Korean peninsula. The abbreviations for each source are explained in Table \ref{tab:data_list}. The blue cells indicate the corpora that we organized and released for the first time under open license terms.}
  \label{fig:teaser}
\end{figure}

\definecolor{level1}{HTML}{FFF8E1}
\definecolor{level2}{HTML}{FFE082}
\definecolor{level3}{HTML}{FFB74D}
\definecolor{level4}{HTML}{D68A54}

\newcommand{\hlone}[1]{\colorbox{level1}{\strut#1}}
\newcommand{\hltwo}[1]{\colorbox{level2}{\strut#1}}
\newcommand{\hlthree}[1]{\colorbox{level3}{\strut#1}}
\newcommand{\hlfour}[1]{\colorbox{level4}{\strut#1}}

\begin{table*}[ht]
  \centering
  \scriptsize
  \setlength{\tabcolsep}{3pt}
  \renewcommand{\arraystretch}{1.0}
  \newcommand{\catfmt}[1]{\textbf{#1}}
  \begin{tabularx}{\textwidth}{
      >{\centering\arraybackslash}m{0.08\textwidth}
      >{\raggedright\arraybackslash}m{0.28\textwidth}
      >{\raggedright\arraybackslash}m{0.58\textwidth}
    }
    \toprule
    \textbf{Period} & \multicolumn{1}{c}{\textbf{Language}} & \multicolumn{1}{c}{\textbf{Text}} \\
    \midrule
    19c-- &
    \catfmt{Modern Korean} (Hangul) &
    {\fontyet{그래도 벌써 몇 년 전 일입니다.}\par\vspace{0pt}\textcolor{gray}{But it was already several years ago.}} \\

    20c-- &
    \catfmt{North Korean} (Hangul) &
    {\fontyet{닭\hlone{료리}는 먹지 않겠어. \hlone{남새료리}를 먹자.}\par\vspace{0pt}\textcolor{gray}{I won't eat chicken dishes. Let's eat vegetable dishes.}} \\

    19c-- &
    \catfmt{Modern Korean} (Hanja-Hangul Mixed Script)   &
    {\fontyet{\hlthree{標}로 풀게된 \hlthree{漢文易解法}}\par\vspace{0pt}\textcolor{gray}{Chart-Based Method for Interpreting Classical Chinese}} \\

    19--20c &
    \catfmt{Japanese} (Colonial Era)    &
    {\fontyet{\hlthree{又大ニ民心}\hlfour{ヲ}\hlthree{收}\hlfour{ムル}\hlthree{ニ似}\hlfour{タリ}}\par\vspace{0pt}\textcolor{gray}{It also seemed that they greatly won the hearts of the people.}} \\

    17--19c &
    \catfmt{Early Modern Korean} (Hanja-Old Hangul Mixed Script) &
    {\fontyet{\hlthree{驚怯逃走}\hltwo{하야}\hlthree{不得禁防}\hltwo{이올두}\hlthree{相考施行伏望}}\par\vspace{0pt}\textcolor{gray}{Frightened, they fled, and prevention was impossible. We respectfully request your examination and action.}} \\

    17--19c &
    \catfmt{Early Modern Korean} (Old Hangul)   &
    {\fontyet{\hltwo{밧바 이만 뎍노라}}\par\vspace{0pt}\textcolor{gray}{I'm busy, so I'll leave it at this.}} \\

    10--16c &
    \catfmt{Middle Korean} (Old Hangul)   &
    {\fontyet{\hltwo{엇디 인다 먼  긔별 도 요이 더옥 듣 디 몯니 념녜 무궁 다}}\par\vspace{0pt}\textcolor{gray}{How are you? As I cannot hear news from afar these days, my worries are endless.}} \\

    7--20c &
    \catfmt{Korean-style Sinitic} (Idu) &
    {\fontyet{\hlthree{此明文內\underline{乙用良}告官辨正\underline{爲乎事}}}\par\vspace{0pt}\textcolor{gray}{The matter is hereby submitted to the authorities for review and determination.}} \\

    --20c &
    \catfmt{Classical Chinese} (Hanja) &
    {\fontyet{\hlthree{八月庚辰 月暈, 內赤中黃, 又與歲星, 同舍.}}\par\vspace{0pt}\textcolor{gray}{On Gyeongjin day in the 8th month, a red and yellow lunar halo appeared with Jupiter.}} \\
    \bottomrule
  \end{tabularx}
  \caption{Textual diversity in the \corpus{}. Selected texts illustrate the linguistic diversity across different eras (periods are approximate). Highlighting distinguishes key scripts and lexical features---such as Hanja, Japanese, Old Hangul, and North Korean vocabulary---from standard Hangul. An English translation is provided for each example.}
  \label{tab:text_example_v2}
\end{table*}

However, computational research on Korean has predominantly focused on modern, Hangul-based texts, largely neglecting the intermediate period of Hanja-Hangul mixed script.
A significant barrier to this research is the lack of accessible historical corpora.
Most corpora created by the Korean government (\eg Sejong Corpus, Modu Corpus) are distributed under restrictive licenses that make releasing derivative works nearly impossible~\cite{hwang2016}.
While some institutions release digitized documents online, they often do not offer them as downloadable datasets.
As a result, interested researchers have been forced to individually crawl numerous websites to compile their own datasets.

To address this problem, we introduce the \corpus{}, a large-scale, openly licensed dataset.
We collected and processed 17.7 million documents from 19 distinct sources, totaling 5.1 billion tokens and covering the period from the 7th century through 2025.
As illustrated in Figure~\ref{fig:teaser}, the collection encompasses six languages: Korean (Middle, Early Modern, Modern, North), Classical Chinese, and Modern Japanese.
This is, to our knowledge, the first openly licensed corpus to offer broad temporal coverage of these historical stages, particularly Middle Korean and Early Modern Korean, as well as under-represented writing systems, including Korean-style Sinitic (Idu) and Hanja-Hangul mixed script.
Table~\ref{tab:text_example_v2} provides selected examples that illustrate this linguistic diversity across different eras.

Using this corpus, we quantitatively analyze major linguistic shifts.
Our analysis of Korean-style Sinitic (Idu) shows its usage peaked in the 1860s before declining sharply, particularly after the 1894 Kabo Reform.
We also trace the diachronic transition from Hanja to Hangul, finding it was a rapid transformation rather than a gradual shift; writing exclusively in Classical Chinese dominated written records until 1890, but by 1980, Hangul comprised over 93\% of characters.
Furthermore, we quantify the lexical divergence of North Korean, which causes modern tokenizers to produce up to 51 times higher out-of-vocabulary (OOV) rates due to its unique orthography for loanwords and distinct native vocabulary.
We release the corpus and our processing code under the CC BY-NC 4.0 and MIT licenses, respectively, to facilitate further research in Korean diachronic linguistics and historical NLP.\footnote{Dataset and code available at \url{https://github.com/seyoungsong/OKHC}.}

\section{Background}
\label{sec:background}
In this section, we examine the historical background of the Korean language and writing system and discuss how they have changed over time.

\subsection{Historical Development of Korean}
\label{sec:historical_development}

The Korean language, with its earliest extant records dating back to the early 5th century~\cite{kim1998variation, bailble2016history}, has undergone substantial changes.
While multiple periodization schemes have been proposed, reflecting different scholarly foci and periodization criteria, we adopt the four-period scheme~\cite{nklc2015korean} for consistency and clarity.
This provides a useful overview by combining dynastic transitions and major linguistic changes as criteria for division.

\paragraph{Ancient (--10C)}
This period spans from before the Common Era to the establishment of the Goryeo dynasty in 918.
During this period, Chinese characters were introduced to the Korean peninsula and became the primary medium of written Korean, known as \textit{Hanja}~\cite{choo2016use}.
Only a few written records from this era have survived.

\paragraph{Medieval (10C--16C)}
The creation of \textit{Hangul}, the native Korean alphabet, in 1443 marked a turning point in the history of the Korean language~\cite{pae2018writing}.
Hangul aimed to mitigate the disparity between native spoken Korean and written Korean borrowed from Chinese characters.
Yet its impact was gradual, as written Korean continued to rely heavily on \textit{Hanja}.
At the same time, the language differed from its modern form, most notably through a tonal system.

\paragraph{Early Modern (17C--late 19C)}
The Korean language underwent significant changes around the 17th century.
In particular, it experienced a gradual loss of its tonal system, including the tonal mark (\textit{pangjeom}, 傍點, ``side marks'') and the vowel (\textit{arae-a}, ㆍ).
In addition, the overall grammar structure became simpler than in earlier periods.
Furthermore, the devastation following the Imjin War (1592--1598) led to the loss of older texts and the weakening of the orthographic conventions that had been maintained in previous eras.

\paragraph{Modern (late 19C--present)}
During this period, Korean underwent dynamic changes shaped by major historical events.
In the late 19th century, Hangul was granted official status as the national script in the wake of the modernization movement.
However, Japanese colonial rule (1910--1945) restricted the use of Korean language and taught Japanese ideology and language~\cite{pak2011assimilation}.
Afterwards, it gradually diverged into distinct varieties in the north and the south, following the division of the peninsula (1945).

\subsection{Diverse Writing Systems in Korean}
\label{sec:script_diversity}

Korean has undergone complex, multi-layered changes in its writing systems.

\paragraph{Korean-style Sinitic}
After the presumed introduction of Chinese characters around the 2nd century BCE~\cite{eom2002origin}, the disparity between spoken Korean and written Chinese motivated the Korean-style Sinitic systems, such as \textit{Idu} (writing Chinese characters in Korean word order) \cite{king2022idu}, \textit{Gugyeol} (adding grammatical markers to Classical Chinese texts) \cite{kim2004history}, and \textit{Hyangchal} (used mainly for transcribing vernacular poetry, \textit{hyangga}) \cite{lee2011history}.

\paragraph{Sino-Korean Mixed Script}
This system refers to the combined use of \textit{Hanja} and \textit{Hangul}, and reflects the coexistence of the two writing systems following the invention of \textit{Hangul} in the 15th century~\cite{pae2018writing}.
This system became widespread in the late 19th century and continued into the modern era~\cite{joyce2019writing}.

\begin{quote}
  \fontyet{第九條, 法律命令은 다 國文으로 本을 삼 漢譯을 附며 或國、漢文을 混用홈。}

  \textcolor{gray}{\small \emph{Article 9: All laws and ordinances shall use the national script as the standard, accompanied by a Chinese translation, or may be written in a mixed script.}}
\end{quote}

\paragraph{Orthographic Divergence after Korean Division}
While both North and South Korea established their orthographic systems based on the Unified Korean Orthography (1933), they have diverged through separate revision processes after the division~\cite{lee2021language_norms}.
Moreover, in South Korea, standardization was primarily based on the Seoul dialect, whereas in North Korea, it was centered on the Pyongyang variety~\cite{song2015language}.

\section{Open Korean Historical Corpus}
\label{sec:data}

This section details the construction of the \corpus{}, outlining the methodology for data collection, text preprocessing, language identification, and schema design.
We provide a statistical overview of the corpus and a discussion of the legal considerations for its release.

\begin{table*}[t]
  \centering
  \resizebox{\textwidth}{!}{%
    \begin{tabular}{@{}llccrrrl@{}}
      \toprule
      \textbf{Abb.} & \textbf{Source}                          & \textbf{Lic.} & \textbf{Pub. Years} & \textbf{Size}  & \textbf{Documents}  & \textbf{Avg. Len.} & \textbf{Languages}                   \\
      \midrule
      NNL  & Naver News Library                       & \tmark                & 1920-1999                  & 13 GB          & 13,536,494          & 385                  & Modern Korean                                    \\
      DRS  & Diaries of the Royal Secretariat         & \cmark                & 1623-1910                  & 750 MB         & 1,792,187           & 165                  & CC, Idu                                           \\
      KLC  & Korean Literary Collections              & \cmark                & 886-1933                   & 656 MB         & 652,405             & 335                  & CC                                           \\
      KCNA & Korean Central News Agency               & \xmark                & 1998-2025                  & 320 MB         & 170,472             & 741                  & North Korean                                     \\
      KNA  & Korean Newspaper Archive                 & \cmark                & 1883-1952                  & 187 MB         & 364,409             & 210                  & Modern Korean                                    \\
      AJD  & Annals of the Joseon Dynasty             & \cmark                & 1392-1928                  & 182 MB         & 413,131             & 173                  & CC                                           \\
      RDR  & The Records of Daily Reflections         & \cmark                & 1760-1910                  & 152 MB         & 338,084             & 153                  & CC                                           \\
      RDGO & Records and Documents of the Government Offices & \cmark                & 1637-1910                  & 130 MB         & 130,143             & 380                  & CC, Modern Korean, Idu                      \\
      MCM  & Modern and Contemporary Magazines        & \cmark                & 1896-1943                  & 121 MB         & 15,326              & 3,228                & Modern Korean                                    \\
      GUM  & GongU Madang                             & \cmark                & 1019-1995                  & 119 MB         & 13,291              & 3,725                & Modern Korean, CC                            \\
      JKU  & Journal of Kim Il Sung University        & \xmark                & 2014-2025                  & 104 MB         & 39,723              & 1,261                & North Korean                                     \\
      AKS  & Academy of Korean Studies                & \cmark                & 695-1985                   & 72 MB          & 55,482              & 502                  & CC, Early Mod Ko, Mid Ko, Idu \\
      RBDC & Records of the Border Defense Council    & \cmark                & 1616-1892                  & 55 MB          & 93,528              & 233                  & CC                                           \\
      RJL  & Records of the Japanese Legation         & \cmark                & 1893-1913                  & 35 MB          & 22,502              & 617                  & Japanese                                         \\
      KIKS & Kyujanggak Institute for Korean Studies  & \cmark                & 1395-1953                  & 23 MB          & 32,487              & 263                  & CC, Idu                                     \\
      HG   & History of Goryeo                        & \cmark                & 1451                       & 4 MB           & 20,047              & 79                   & CC                                           \\
      NHMA & National Hangeul Museum Archive          & \cmark                & 1628-1988                  & 3 MB           & 1,669               & 743                  & Modern Korean, Early Mod Ko               \\
      LKU  & Literary Works of Kim Il Sung University & \xmark                & 2015-2025                  & 1 MB           & 274                 & 1,759                & North Korean                                     \\
      HTK  & History of the Three Kingdoms            & \cmark                & 1145                       & 667 KB         & 4,613               & 58                   & CC                                           \\
      \midrule %
      \textbf{Total} &                          & \textbf{}        & \textbf{695-2025}          & \textbf{16 GB} & \textbf{17,696,267} & \textbf{357}         & \textbf{}                                        \\
      \bottomrule
    \end{tabular}
  }
  \caption{Overview of the data sources and statistics. The Lic. column indicates accessibility: \cmark{} for public domain or openly licensed; \tmark{} for partially restricted sources where texts published before 1963 are public domain; and \xmark{} for copyrighted materials, for which only metadata and links are provided. Avg. Len. denotes the average document length in characters. In the Languages column, CC, Early Mod Ko, and Mid Ko refer to Classical Chinese, Early Modern Korean and Middle Korean, respectively. Idu, technically a writing system rather than a distinct language, is also listed in this column.}
  \label{tab:data_list}
\end{table*}

\subsection{Data Collection}
\label{sec:collection}

To build a comprehensive resource, we target key languages and writing systems with historical significance on the Korean peninsula.
The languages include several varieties of Korean (Middle, Early Modern, Modern, and North Korean), Classical Chinese, and Japanese.
The inclusion of Japanese addresses its enforced use during the colonial period (1910--1945), which left a substantial body of written records and influenced the Korean language post-independence.
The corpus covers a range of writing systems, including Korean-style Sinitic (Idu), Hanja-Hangul mixed script, Old Hangul, modern Hangul, and the Japanese writing system.
Sources are selected based on their ability to provide digitized original texts accompanied by publication year metadata.
Data is gathered primarily by web scraping institutional websites using Python libraries such as BeautifulSoup, HTTPX, and Selenium, and by direct downloads where available.
Data collection occurred from May 2025 to October 2025, with the most recent documents from North Korean news sources dated June 19, 2025.

\paragraph{Attribution.}

The corpus is composed of materials from numerous institutions, as detailed in Table~\ref{tab:data_list}.
A significant portion of the historical documents is sourced from the National Institute of Korean History (NIKH)\footnote{\url{https://db.history.go.kr}}, which provides the \emph{Diaries of the Royal Secretariat}\footnote{\url{https://sjw.history.go.kr}}, the \emph{Annals of the Joseon Dynasty}\footnote{\url{https://sillok.history.go.kr}}, \emph{Records and Documents of the Government Offices}, \emph{Modern and Contemporary Magazines}, the \emph{Records of the Border Defense Council}, the \emph{Records of the Japanese Legation}, the \emph{History of Goryeo}, and the \emph{History of the Three Kingdoms}.
Additional public archives include the National Library of Korea (NLK)\footnote{\url{https://www.nl.go.kr}}, which contributes the \emph{Korean Newspaper Archive}, and the National Hangeul Museum\footnote{\url{https://archives.hangeul.go.kr}}, which supplies its \emph{Archive}.
The Institute for the Translation of Korean Classics (ITKC)\footnote{\url{https://db.itkc.or.kr}} provides the \emph{Korean Literary Collections}, while the Kyujanggak Institute for Korean Studies\footnote{\url{https://kyudb.snu.ac.kr}} offers \emph{The Records of Daily Reflections} and its collection of old documents.
The Academy of Korean Studies (AKS)\footnote{\url{https://www.aks.ac.kr}} contributes a range of materials, including old Korean books, document collections, royal court documents, and Hangul letters, which we aggregate from four of its distinct web archives.
To include North Korean texts, data is sourced from Kim Il Sung University\footnote{\url{http://www.ryongnamsan.edu.kp}} (\emph{Journal} and \emph{Literary Works}) and the Korea News Service\footnote{\url{http://www.kcna.co.jp}}, which archives news from the \emph{Korean Central News Agency (KCNA)}.
The corpus is further supplemented by materials from the Korea Copyright Commission's \emph{GongU Madang}\footnote{\url{https://gongu.copyright.or.kr}}, which span a wide historical period, and by modern news articles from five major newspapers (\emph{Kyunghyang Shinmun, Maeil Business Newspaper, The Chosun Ilbo, The Dong-A Ilbo, and The Hankyoreh}) provided by Naver\footnote{\url{https://newslibrary.naver.com}}.

\subsection{Text Preprocessing}
\label{sec:preprocessing}

Our preprocessing pipeline first applied universal normalization to all documents and then used source-aware heuristics to remove archival and digitization artifacts.
The initial normalization stage standardized Unicode text to its NFKC form to resolve compatibility variants, collapsed irregular whitespace, and removed non-printable control characters.
The subsequent cleaning stage applied targeted rules to specific sources.
These operations included:

\begin{itemize}
  \item Removing boilerplate text that indicates missing content or metadata.
  \item Removing modern in-line translations and isolating original texts.
  \item Stripping erroneous metadata, like modern Korean titles or image captions that were merged into the main text field.
  \item Filtering out documents with a high ratio of noise-to-text, which often resulted from encoding errors.
\end{itemize}

\subsection{Language and Script Identification}
\label{sec:langid}

We perform an initial analysis using two methods: a GlotLID~\cite{kargaran-etal-2023-glotlid} language identification model constrained to predict from a small set of relevant languages (\eg Korean, Japanese, Chinese), and a character-level script analysis that computes the proportion of Hangul, Hanja, and Kana.
The final classification is determined by a set of rules that integrate these preliminary results with source-specific metadata.
This heuristic approach is necessary, as modern language identification models are less reliable for historical texts.
For instance, documents from sources known to contain only Classical Chinese, such as the \emph{Annals of the Joseon Dynasty}, are categorized as Classical Chinese based on their origin.
For Korean-language texts, we assign a historical period---Middle, Early Modern, or Modern Korean---based on the document's publication year.

\subsection{Data Schema and Format}
\label{sec:metadata}

\begin{figure}[t]
  \centering
  \includegraphics[width=\columnwidth]{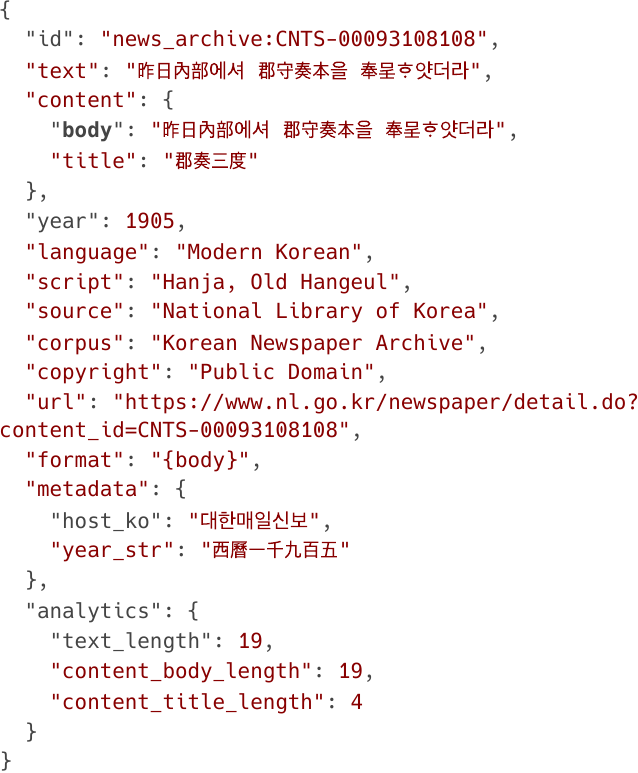}
  \caption{An example of a JSON data instance from the \corpus{}.}
  \label{fig:json_sample}
\end{figure}

The corpus is distributed in the JSON Lines format, where each line is a self-contained JSON object representing a single document.
As shown in Table~\ref{tab:json_schema}, the schema is designed for flexibility and traceability, allowing for the aggregation of documents from diverse sources.
Each entry includes fields for identification and attribution, manually curated classification labels for targeted analysis, and distinct fields for both raw and normalized text to ensure data integrity and usability for NLP research.
An example of a single data instance in JSON format is shown in Figure~\ref{fig:json_sample}.

\begin{table}[t]
  \centering
  \small
  \begin{tabularx}{\linewidth}{@{} l l X @{}}
    \toprule
    \textbf{Key} & \textbf{Type} & \textbf{Description} \\
    \midrule
    id & string & Unique document identifier. \\
    text & string & Main content of the document. \\
    content & object & Raw textual components. \\
    year & integer & Publication year of the document. \\
    language & string & Primary language of the text. \\
    script & string & Primary script used in the text. \\
    source & string & Source institution or archive. \\
    corpus & string & The name of the collection. \\
    copyright & string & Copyright or license status. \\
    url & string & Link to the original document. \\
    format & string & Template for `text' field. \\
    metadata & object & Object for source-specific fields. \\
    analytics & object & Computed text metrics. \\
    \bottomrule
  \end{tabularx}%
  \caption{JSON Schema. The schema provides fields for identification, classification, text content, and metadata.}
  \label{tab:json_schema}
\end{table}

\subsection{Corpus Statistics and Distribution}
\label{sec:statistics}

\begin{figure}[t]
  \centering
  \includegraphics[width=\linewidth]{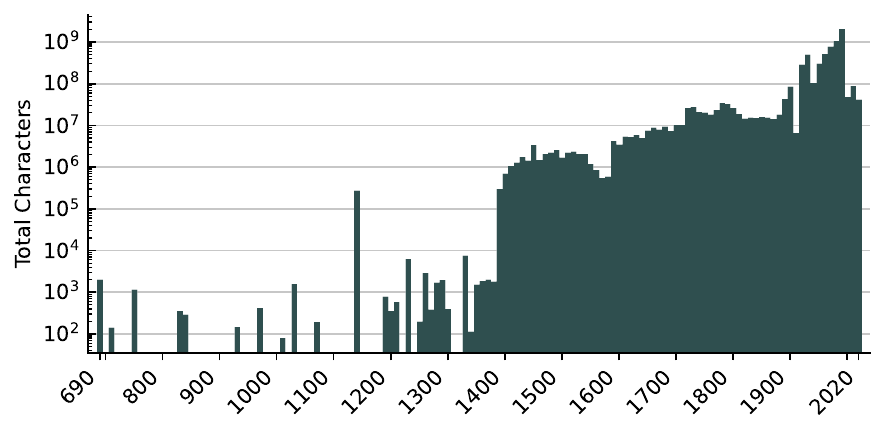}
  \caption{Temporal distribution of the corpus. The total number of characters per decade is plotted on a logarithmic scale. The data volume increases significantly following the establishment of the Joseon dynasty in 1392 and shows a consistent upward trend into the modern era.}
  \label{fig:token_stack}
\end{figure}

The \corpus{} is a large-scale collection of 17.7 million documents from the 19 sources detailed in Table~\ref{tab:data_list}.
It contains 6.3 billion characters, which corresponds to 5.1 billion tokens as measured by the \texttt{o200k\_base} tokenizer\footnote{\url{https://github.com/openai/tiktoken}}, making it a substantial resource for the historical study of the Korean language.
The corpus's temporal distribution, visualized in Figure~\ref{fig:token_stack}, shows a heavy concentration in later periods.
The apparent data sparsity before 1392 is partly due to visualization constraints, as many individual historical documents, particularly from non-periodical sources like AKS, KLC, and NHMA, lack precise year-level metadata and are therefore not plotted.
The data volume increases substantially with the establishment of the Joseon Dynasty in 1392, a period known for systematic record-keeping, and then expands exponentially from the late 19th century with the advent of mass media.
This results in a sampling bias where modern texts are far more represented than those from earlier eras.

\subsection{Legal Considerations}
\label{sec:legal}

This corpus is distributed under the Creative Commons Attribution-NonCommercial 4.0 International license in compliance with South Korean copyright law, \emph{sui generis} database rights, and the National Security Act.
Texts published before 1963, which are in the public domain, are included in full, accounting for 41.9\% of all documents and 2.1 billion tokens.
For copyrighted materials, such as post-1963 articles or modern translations, we provide the title, metadata, and direct URLs to the original sources to avoid infringement.
A detailed breakdown of licensing and distribution formats is provided in Appendix~\ref{sec:data_availability}.
The corpus's non-commercial academic and research license explicitly falls within the permissible use exceptions for database rights, allowing reproduction and distribution for educational and scholarly purposes without commercial intent.
North Korean texts are also provided as title, URLs, and metadata; their inclusion is strictly for linguistic and scholarly analysis, a non-ideological purpose that complies with the National Security Act.

\section{Discussion}
\label{sec:discussion}
\subsection{Temporal Dynamics of Korean-Style Sinitic}
\label{sec:ksc_id}

\begin{figure}[b]
  \centering
  \includegraphics[width=\columnwidth]{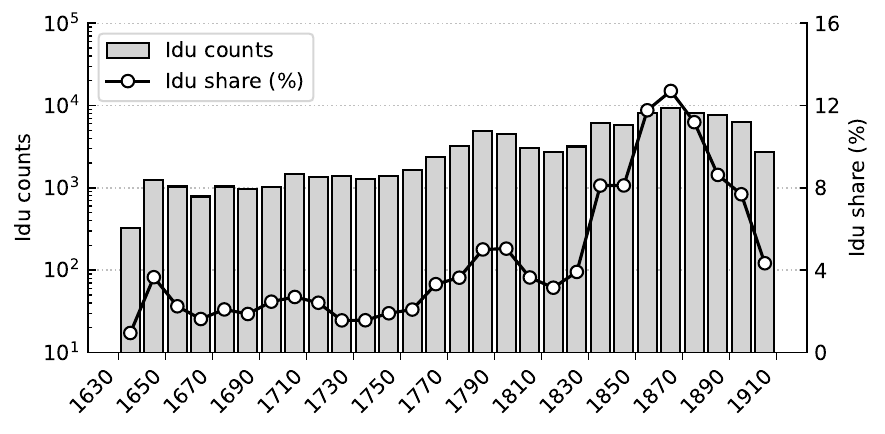}
  \caption{Temporal distribution of Korean-style Sinitic (Idu) usage (1630s-1910s). The bars (left y-axis, logarithmic scale) represent the total count of Idu instances per decade. The black line with circular markers (right y-axis, percentage) illustrates the share of Idu within the analyzed documents from the 1630s to the 1900s.}
  \label{fig:ksc}
\end{figure}

To quantitatively analyze the temporal distribution of Korean-style Sinitic (Idu), we first compiled a merged lexicon from two Idu dictionaries~\cite{dku_idu_dictionary, aks_idu_examples}.
Using this lexicon, we identify Idu documents using a longest non-overlapping matching method with the Aho--Corasick algorithm~\cite{aho1975efficient}.
Documents are scored by the frequency of Idu markers, normalized by the Hanja character count.
A document is classified as Idu if this score exceeds length-stratified thresholds (1.04\% for $\le$100 chars; 0.86\% for 101--300 chars; 0.38\% for $\ge$301 chars); exclusion lists are employed to filter borderline cases and minimize false positives.
To mitigate proportional bias, our analysis is restricted to the 1637--1910 period, which ensures comparable coverage across the four subcorpora (RDGO, DRS, AKS, KIKS) of administrative and legal records; we also exclude sources with an Idu share below 1\%.

As shown in Figure~\ref{fig:ksc}, the data reveal a gradual increase in Idu usage from 1637 to the 1860s, consistent with higher survival rates for later-period documents~\cite{lee2004old-documents}, followed by a steady decline through 1910.
The post-1860s decline indicates factors beyond preservation artifacts.
The most significant drop occurs between the 1890s and 1900s, strongly correlating with major language policy changes: the Kabo Reform (1894), which mandated Hangul in official documents, and a 1908 directive favoring Hanja--Hangul script over Idu.
This temporal alignment suggests language policy accelerated the decline of Idu in administrative contexts.
We interpret these figures as conservative lower bounds, as the corpus includes only digitized materials and our classification method was optimized for high precision.

\subsection{The Diachronic Shift from Hanja to Hangul}
\label{sec:script_change}

\begin{figure}[b]
  \centering
  \includegraphics[width=\linewidth]{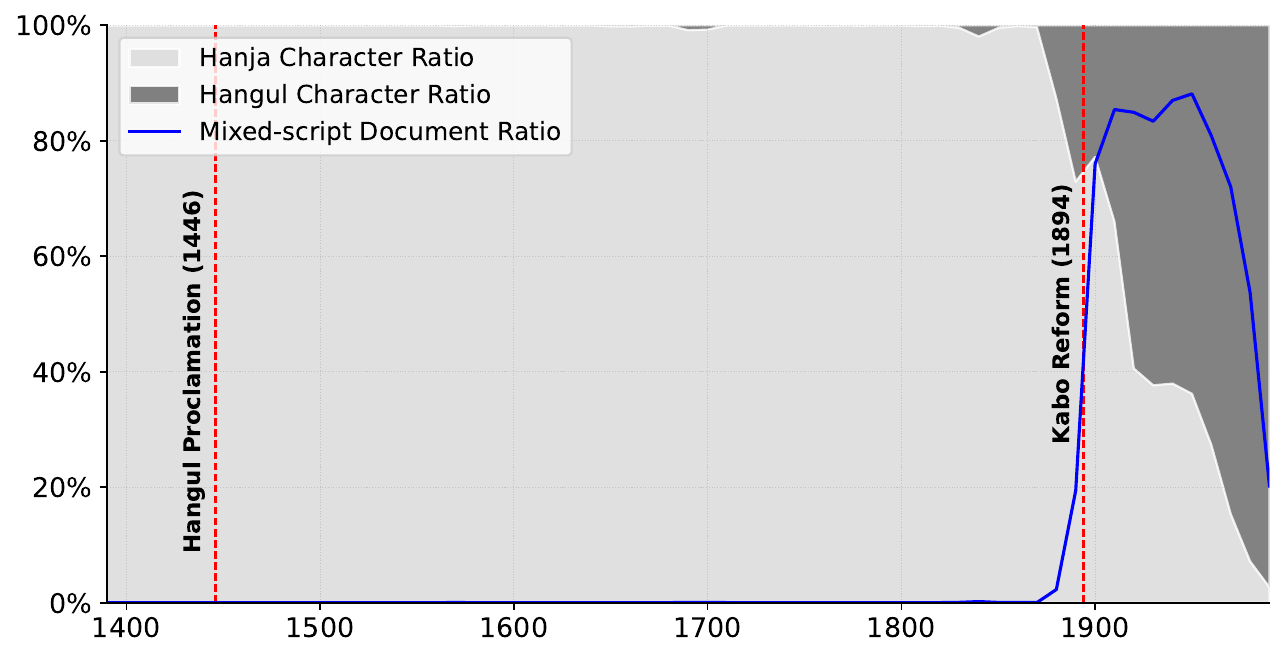}
  \caption{Script transition from Hanja to Hangul between 1390 and 1990, aggregated by decade. The stacked area chart shows the proportion of Hanja (light gray) versus Hangul (dark gray) characters. The blue line indicates the ratio of documents written in Hanja-Hangul mixed script.}
  \label{fig:script_change}
\end{figure}

\begin{table*}[t]
  \centering
  \small
  \begin{tabularx}{\textwidth}{@{} l X @{}}
    \toprule
    \textbf{Source} & \textbf{OOV Words} \\
    \midrule
    NK News & {\textbf 돐}에, 파{\textbf 쑈}적인, {\textbf 뻬}루, 도이{\textbf 췰}란드, {\textbf 뷸}레찐은, 도{\textbf 꾜}, {\textbf 윁}남사회주의공화국, {\textbf 줴}쳤다, 브라질주체사상연구{\textbf 쎈}터, {\textbf 뽈}스까, {\textbf 먄}마, 에{\textbf 꽈}도르, {\textbf 죤}, {\textbf 홰}불, {\textbf 챠}베스, 마{\textbf 쩨}고라, {\textbf 낟}알털기를, 바{\textbf 씰}레, 그{\textbf 쯘}히, {\textbf 쟝}, 아제르바이{\textbf 쟌}공화국, {\textbf 꼰}스딴찐, 브류{\textbf 쎌}에서, 어깨{\textbf 겯}고, {\textbf 멎}은, {\textbf 겡}이찌, {\textbf 꿩}, {\textbf 윅}또르, {\textbf 쑬}레이만, 세이{\textbf 쉘}공화국\newline{}\vspace{0pt}\textcolor{gray}{anniversary, fascist, Peru, Germany, bulletin, Tokyo, Socialist Republic of Vietnam, shouted, Brazil Juche Idea Study Center, Poland, Myanmar, Ecuador, John, torch, Chavez, Matsegora, grain threshing, Vasile, to that extent, Jean, Republic of Azerbaijan, Constantine, in Brussels, side-by-side, stopped, Genichi, pheasant, Viktor, Suleiman, Republic of Seychelles} \\
    \addlinespace
    SK Web & 가리{\textbf 킵}니다, {\textbf 옅}은, {\textbf 챗}봇, {\textbf 욥}의, {\textbf 큽}니다, {\textbf 퀀}텀, 어{\textbf 댑}터, {\textbf 퀵}, 몽{\textbf 쉘}이의, 배{\textbf 웁}니다, {\textbf 홋}카이도, {\textbf 찝}찝한, 느{\textbf 낍}니다, 왓{\textbf 챠}, 긍{\textbf 휼}을, 보살{\textbf 핌}을, {\textbf 멱}살을, {\textbf 엡}, {\textbf 훅}, {\textbf 옻}칠, {\textbf 헹}궈, 슬{\textbf 펐}다, 비{\textbf 젼}을, {\textbf 켤}레, {\textbf 쯧}, {\textbf 갭}, {\textbf 챕}터, {\textbf 펭}귄, {\textbf 쇳}소리, 만{\textbf 듦}\newline{}\vspace{0pt}\textcolor{gray}{points to, light, chatbot, Job's, is big, quantum, adapter, quick, Mongcael's, learns, Hokkaido, uncomfortable, feels, Watcha, mercy, care, collar, app, hook, lacquer, rinse, was sad, vision, pair, tsk, gap, chapter, penguin, metallic sound, making} \\
    \bottomrule
  \end{tabularx}
  \caption{Top 30 most frequent OOV word stems from the NK News and SK Web corpora, as identified by the KLUE-BERT model. The table displays the most frequent example word that corresponds to each stem (in bold), followed by its English translation.}
  \label{tab:nk_oov_words}
\end{table*}

\begin{table}[t]
  \centering
  \small
  \begin{tabular}{@{}lcrr@{}}
    \toprule
    \textbf{Model} & \textbf{Type} & \textbf{NK News} & \textbf{SK Web} \\ \midrule
    mBERT          & M & 0.2823           & 0.3196          \\
    KLUE-BERT      & K & 0.2035           & 0.0650          \\
    KcBERT         & K & 0.1226           & 0.0024          \\
    XLM-R          & M & 0.0499           & 0.0125          \\ \bottomrule
  \end{tabular}
  \caption{Out-of-Vocabulary rates for North Korean (NK) and South Korean (SK) text. The values represent the percentage of unknown tokens generated by four different encoder models when tokenizing the NK News and SK Web corpora. Type indicates if the model is trained on multilingual (M) or South Korean (K) texts.}
  \label{tab:nk_oov}
\end{table}

To measure the diachronic shift of script usage from Hanja to Hangul, we analyzed all documents in our corpus written in the Korean language or Classical Chinese from 1390 to 1999.
Using regular expressions based on Unicode properties and blocks, we counted Hanja and Hangul characters in each document.
These counts were then aggregated into a timeframe of 10-year decades, and we calculated both the character-level script proportion and the ratio of documents written in the Hanja-Hangul mixed script system, where the threshold is over 10\% of both Hanja and Hangul usage.

As shown in Figure~\ref{fig:script_change}, the data reveals a rapid transformation rather than a gradual shift.
From the release of Hangul in the mid-15th century to the late 19th century, Hanja usage was dominant, with almost all documents written exclusively in Classical Chinese.
A sharp decline in Hanja-exclusive texts began around 1890, which can be interpreted as a result of modernizing efforts such as the Kabo Reform (1894), which mandated Hangul for official documents.
Meanwhile, as Hanja-Hangul mixed script usage rose and then fell, the usage of Hangul increased rapidly.
By the 1980s, Hangul became the primary script, accounting for over 93\% of characters and reflecting language policies that relegated Hanja to a supplementary role.
This result accurately reflects known turning points of Korean language history, and shows that our corpus is useful for such diachronic analysis.

\subsection{A Preliminary Analysis of Tokenizer Coverage on North Korean Text}
\label{sec:nk_analysis}

Decades of political separation have caused the Korean language to diverge lexically between the North and South.
As a preliminary step to understand potential model limitations, we assess how well the tokenizer, the initial stage of language modeling, handles this divergence.
We quantify this gap by measuring the out-of-vocabulary (OOV) rate---the proportion of unknown tokens generated by the tokenizers of four pre-trained encoder models\footnote{Model identifiers are \texttt{google-bert/bert-base-multilingual-cased}, \texttt{FacebookAI/xlm-roberta-base}, \texttt{klue/bert-base}, and \texttt{beomi/kcbert-base}.}---when processing our North Korean news corpus (KCNA) and a comparably-sized South Korean web text baseline.\footnote{The South Korean baseline was a 5\% random sample of the \texttt{HAERAE-HUB/KOREAN-WEBTEXT} corpus, which is available on Hugging Face. Our KCNA corpus contains approximately 170k documents and 126M characters; the sampled SK corpus contains approximately 64k documents and 175M characters.} All texts were pre-processed using NFKC normalization and filtered to retain only complete Hangul syllables.

Our results confirm a measurable lexical gap at the token level.
As shown in Table~\ref{tab:nk_oov}, OOV rates for North Korean text were significantly higher for most models, ranging from 3 to 51 times, though the mBERT model showed a comparable rate.
However, the absolute rates remained low (under 0.3\%), suggesting modern subword tokenizers are surprisingly resilient in segmenting the text.
An analysis of the most frequent OOV tokens (Table~\ref{tab:nk_oov_words}) reveals the divergence stems primarily from North Korea's unique orthography for loanwords (\eg{} \textit{도이췰란드} for Germany) and distinct vocabulary (\eg{} \textit{돐} for anniversary).
This lexical gap, though small, can still disproportionately impact sensitive tasks like named entity recognition, highlighting a clear opportunity for model improvement through vocabulary expansion.

Several limitations apply to this analysis.
The two corpora differ in domain: North Korean news articles versus South Korean general web text, which means topic and genre differences may inflate OOV rates beyond what is attributable to linguistic divergence alone.
Furthermore, North Korean state media is narrow in both topical coverage and stylistic diversity, and may not represent everyday North Korean language use.
The reported OOV rates should therefore be interpreted as an upper bound reflecting both true lexical divergence and domain mismatch.

\section{Related Work}
\label{sec:related_work}
\paragraph{Korean Corpora for Modern NLP}
Large-scale multilingual corpora have incorporated Korean data, but their coverage remains limited in both temporal and linguistic diversity, failing to capture earlier stages or varieties of the language that are essential for studying linguistic change and variation.
Resources such as OSCAR \cite{suarez2019asynchronous}, mC4 \cite{xue-etal-2021-mt5}, and CulturaX \cite{nguyen2023culturax} include Korean data, but these materials primarily reflect modern standardized usage.
Dedicated Korean corpora, including AI-Hub\footnote{\url{https://www.aihub.or.kr}} \cite{nia2020aihub}, koTenTen \cite{koTenTen2018}, Korpora \cite{ko-nlp2019korpora}, and the Open Korean Corpora \cite{cho-etal-2020-open}, have established valuable resources for modern Korean NLP.
However, their scope likewise remains confined to contemporary language.
Moreover, access to AI-Hub datasets is limited to domestic applicants, and their overseas use---including by Korean nationals abroad---is either restricted or requires separate agreements with the responsible institutions and government bodies, limiting global research use.
Overall, existing Korean corpora lack temporal depth and script variation, leaving historical materials and mixed-script texts largely unrepresented.

\paragraph{Historical Korean Corpora}
Historical corpora play a crucial role in preserving cultural heritage and enabling longitudinal analyses of language and society.
Historical corpora initiatives such as Shamela \cite{belinkov2016shamela}, CHisIEC \cite{tang2024chisiec}, and CCOHA \cite{alatrash2020ccoha} illustrate how historical datasets can mitigate temporal bias and support research on long-term linguistic and cultural change.

In the Korean context, the inclusion of classical texts written in Hanja has long been regarded as essential, since Hanja served as the principal written medium prior to the invention and popularization of Hangul.
Several efforts have been made to build historical corpora, including the Integrated Database of Korean Classics \cite{itkc_database} and KoHiCo \cite{kohico}.
The lack of standardized formats and limited script coverage, however, still pose challenges for researchers and hinder reproducibility.

National projects further contribute by encompassing both classical and modern data, but their main challenge lies in restricted accessibility and licensing.
The Sejong Corpus (1998--2007) consists of about 200M tokens across written, spoken, parallel (Korean--English/Japanese), and historical data \cite{An2025everyone}.
However, some obstacles, such as its limited distribution (mainly in DVD format), have hindered its usage.
The Modu Corpus \cite{nikl2019modu} provides diverse datasets (sentiment, NER, parallel texts, newspapers, everyday dialogues) and serves as a de facto standard resource.
However, access requires registration and approval, and its diachronic coverage remains narrow.

To overcome limited accessibility, openness, and script coverage of the existing historical corpora, we introduce an openly available corpus spanning a broad temporal and linguistic range---including \textit{Hanja}, Korean-style Sinitic (\textit{Idu}), Middle and Early Modern Korean, and Japanese texts from the colonial period---integrated into a unified, NLP-ready format for large-scale exploration of Korean language and culture across time.

\section{Conclusion}
\label{sec:conclusion}

In this work, we introduced the \corpus{}, a large-scale, openly licensed dataset of 17.7 million documents designed to address the critical lack of accessible historical data for Korean NLP.
By compiling and standardizing texts from the 7th century to the present, our corpus provides the first comprehensive resource for studying the language's diachronic evolution across diverse languages and writing systems, including Classical Chinese, Korean-style Sinitic, and Hanja-Hangul mixed script.
We present quantitative analyses of notable linguistic changes, including the temporal dynamics of Korean-style Sinitic, the Hanja-to-Hangul transition, and the lexical divergence of North Korean.

Beyond these analyses, the corpus opens up practical possibilities in an area where current Korean NLP tools fall short: they are predominantly trained on modern Hangul and perform poorly on historical texts.
Our resource could serve as a foundation for closing this disconnect --- for instance, by training tokenizers and language models that are robust to diachronic variation in script and orthography, pre-training specialized encoder models for downstream digital humanities tasks on archival materials, or continually pre-training large language models to improve their handling of historical writing systems.
More broadly, the corpus's temporal depth can support historical linguistics research by enabling fine-grained diachronic analyses across genres, regions, and document types.
We release the corpus and our processing code to facilitate such research and to lay the groundwork for making centuries of historical documents more accessible.

\section*{Acknowledgements}
\label{sec:ack}

This research was conducted as part of the Sovereign AI Foundation Model Project(GPU Track), organized by the Ministry of Science and ICT(MSIT) and supported by the National IT Industry Promotion Agency(NIPA), S.Korea. (PJT-26-010017)

This research was supported by the MSIT(Ministry of Science, ICT), Korea, under the Top-Tier AI Global HRD invitation program (RS-2025-25461932) supervised by the IITP(Institute for Information \& Communications Technology Planning \& Evaluation).

This work was supported by the Institute of Information \& Communications Technology Planning \& Evaluation (IITP) with a grant funded by the Ministry of Science and ICT (MSIT) of the Republic of Korea in connection with the Global AI Frontier Lab International Collaborative Research.

\section*{Ethics Statement}
\label{sec:ethics}

The \corpus{} is provided for research in historical linguistics and natural language processing.
To ensure academic integrity, we provide detailed source information for all texts to properly attribute the work of the original creators and digitizing institutions.

We caution users to handle the data with care, as it contains sensitive materials rooted in outdated cultural norms, including gender, religious, class, and regional discrimination.
We have included North Korean texts from state media for linguistic analysis only; this inclusion does not constitute an endorsement of their ideological content, which features political propaganda, censorship, extreme language, and a cult of personality.
We also acknowledge that historical documents contain biases that language models trained on this corpus may learn and reproduce.
To prevent the downstream harms of naive or blind usage, researchers must actively consider and mitigate these biases during model training.

\section*{Limitations}
\label{sec:limits}

The primary limitation of this corpus is its inherent sampling bias.
It is constructed from extant and digitized materials, which do not represent a complete or balanced sample of all historical documents ever produced.
This bias is evident in the temporal distribution of the data, where modern periods are heavily overrepresented compared to earlier eras from which fewer texts have survived or been digitized.
Furthermore, the scope of the corpus is intentionally limited to written texts.
It does not include transcribed spoken language, making it unsuitable for certain linguistic studies, such as those focusing on phonological or conversational phenomena.

\nocite{*}
\section*{Bibliographical References}
\label{sec:reference}
\bibliographystyle{lrec2026-natbib}
\bibliography{references/lrec2026-example}

\clearpage
\onecolumn
\appendix
\section*{Appendix}
\label{sec:appendix}
\section{Data Availability and Licensing}
\label{sec:data_availability}

\begin{table*}[htb]
  \centering
  \resizebox{\textwidth}{!}{%
    \begin{tabular}{@{}lrrrrrr@{}}
      \toprule
      \multirow{2}{*}{\textbf{Source}} & \multicolumn{4}{c}{\textbf{License}} & \multicolumn{2}{c}{\textbf{Distribution}} \\
      & \textbf{All Rights Reserved} & \textbf{Public Domain} & \textbf{CC BY-NC-ND 2.0 KR} & \textbf{KOGL Type 1} & \textbf{Text \& Metadata} & \textbf{Metadata Only} \\ \midrule
      NNL & 10,063,510 & 3,472,984 & 0 & 0 & 25.7\% & 74.3\% \\
      DRS & 0 & 1,792,187 & 0 & 0 & 100.0\% & 0.0\% \\
      KLC & 0 & 652,405 & 0 & 0 & 100.0\% & 0.0\% \\
      KCNA & 170,472 & 0 & 0 & 0 & 0.0\% & 100.0\% \\
      KNA & 0 & 364,409 & 0 & 0 & 100.0\% & 0.0\% \\
      AJD & 0 & 413,131 & 0 & 0 & 100.0\% & 0.0\% \\
      RDR & 0 & 338,084 & 0 & 0 & 100.0\% & 0.0\% \\
      RDGO & 0 & 130,143 & 0 & 0 & 100.0\% & 0.0\% \\
      MCM & 0 & 15,326 & 0 & 0 & 100.0\% & 0.0\% \\
      GUM & 0 & 13,291 & 0 & 0 & 100.0\% & 0.0\% \\
      JKU & 39,723 & 0 & 0 & 0 & 0.0\% & 100.0\% \\
      AKS & 0 & 55,409 & 73 & 0 & 100.0\% & 0.0\% \\
      RBDC & 0 & 93,528 & 0 & 0 & 100.0\% & 0.0\% \\
      RJL & 0 & 22,502 & 0 & 0 & 100.0\% & 0.0\% \\
      KIKS & 0 & 32,487 & 0 & 0 & 100.0\% & 0.0\% \\
      HG & 0 & 20,047 & 0 & 0 & 100.0\% & 0.0\% \\
      NHMA & 0 & 1,654 & 0 & 15 & 100.0\% & 0.0\% \\
      LKU & 274 & 0 & 0 & 0 & 0.0\% & 100.0\% \\
      HTK & 0 & 4,613 & 0 & 0 & 100.0\% & 0.0\% \\
      \textbf{Total} & \textbf{10,273,979} & \textbf{7,422,200} & \textbf{73} & \textbf{15} & \textbf{41.9\%} & \textbf{58.1\%} \\ \bottomrule
    \end{tabular}%
  }
  \caption{Legal licensing and distribution formats across the 19 sources in the \corpus{}. Documents are categorized into four license types, which dictate their available distribution format. Full text is provided for openly licensed materials (Public Domain, CC BY-NC-ND 2.0 KR, KOGL Type 1), while copyrighted sources (All Rights Reserved) are restricted to metadata and URLs.}
  \label{tab:data_list_copyright}
\end{table*}

Table~\ref{tab:data_list_copyright} presents the licensing and distribution format for all 19 sources. Overall, 41.9\% of documents have full text included, while 58.1\% are metadata-only, containing title, year, language, and a URL to the original source. The metadata-only category consists of copyrighted (``All Rights Reserved'') materials for which we lack redistribution rights. The vast majority of these---10.1 million out of 10.3 million---come from the Naver News Library (post-1963 newspaper articles). The remaining copyrighted entries are the three North Korean sources (KCNA, JKU, LKU), for which metadata-only distribution also ensures compliance with the National Security Act.

Among openly licensed materials, most documents are in the public domain. Two additional license types appear in small quantities: 73 documents from AKS under CC BY-NC-ND 2.0 KR, a non-commercial, no-derivatives license; and 15 documents from NHMA under KOGL (Korea Open Government License) Type 1\footnote{\url{https://www.kogl.or.kr/info/license.do}}, an attribution-only public license issued by the South Korean government that is functionally equivalent to CC BY but not officially declared interoperable with Creative Commons.

\end{document}